\documentclass{vgtc}                          % final (conference style)
\ifpdf%                                % if we use pdflatex
  \pdfoutput=1\relax                   % create PDFs from pdfLaTeX
  \usepackage{graphicx}                % allow us to embed graphics files
  \DeclareGraphicsExtensions{.pdf,.png,.jpg,.jpeg} % for pdflatex we expect .pdf, .png, or .jpg files
\else%                                 % else we use pure latex
  \usepackage{graphicx}                % allow us to embed graphics files
  \DeclareGraphicsExtensions{.eps}     % for pure latex we expect eps files
\fi%

\graphicspath{{figures/}{pictures/}{images/}{./}} % where to search for the images

\usepackage{microtype}                 % use micro-typography (slightly more compact, better to read)
\PassOptionsToPackage{warn}{textcomp}  % to address font issues with \textrightarrow
\usepackage{textcomp}                  % use better special symbols
\usepackage{mathptmx}                  % use matching math font
\usepackage{times}                     % we use Times as the main font
\usepackage{cite}                      % needed to automatically sort the references
\usepackage{tabu}                      % only used for the table example
\usepackage{booktabs}                  % only used for the table example
\usepackage{cancel}
\usepackage{ulem}
\onlineid{1973}

\vgtccategory{Research}

\vgtcinsertpkg

\title{EmoFace: Audio-driven Emotional 3D Face Animation}

\author{
Chang Liu \thanks{e-mail: frenkiedejong@sjtu.edu.cn}\\ %
     \scriptsize Shanghai Jiao Tong University%
\and Qunfen Lin \thanks{e-mail: volleylin@tencent.com}\\ %
     \scriptsize Tencent Games%
\and Zijiao Zeng\thanks{e-mail: zijiaozeng@tencent.com}\\ %
     \scriptsize Tencent Games
\and Ye Pan \thanks{e-mail: whitneypanye@sjtu.edu.cn (Corresponding author)} \\ %
     \scriptsize Shanghai Jiao Tong University
}

\teaser{
  \centering
  \includegraphics[width=\linewidth]{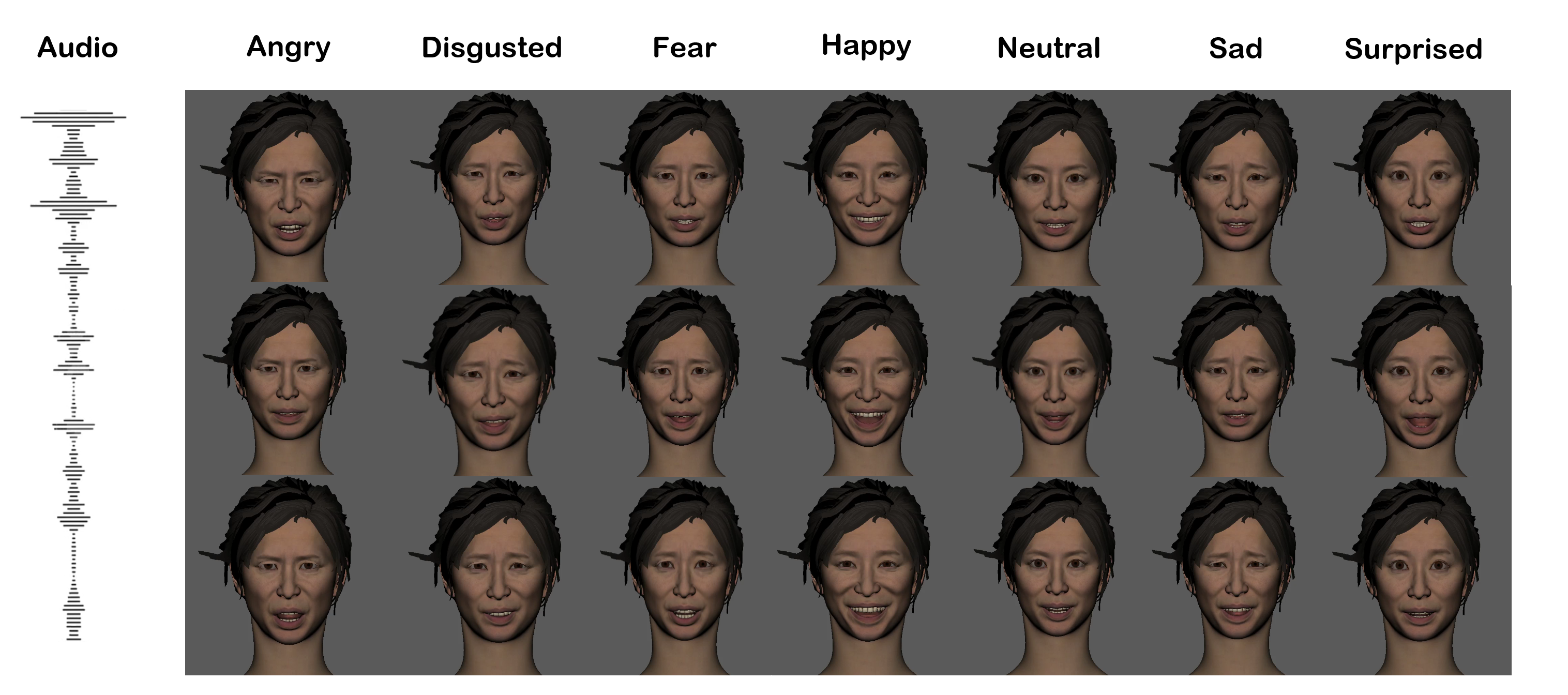}
  \caption{Given an audio clip and a target emotion, EmoFace can generate talking heads with fully controllable emotions. This figure shows generated facial animation using the same audio and different emotions.}
  \label{fig:teaser}
}

\abstract{
Audio-driven emotional 3D face animation aims to generate emotionally expressive talking heads with synchronized lip movements. However, previous research has often overlooked the influence of diverse emotions on facial expressions or proved unsuitable for driving MetaHuman models. In response to this deficiency, we introduce EmoFace, a novel audio-driven methodology for creating facial animations with vivid emotional dynamics. Our approach can generate facial expressions with multiple emotions, and has the ability to generate random yet natural blinks and eye movements, while maintaining accurate lip synchronization. We propose independent speech encoders and emotion encoders to learn the relationship between audio, emotion and corresponding facial controller rigs, and finally map into the sequence of controller values. Additionally, we introduce two post-processing techniques dedicated to enhancing the authenticity of the animation, particularly in blinks and eye movements. Furthermore,  recognizing the scarcity of emotional audio-visual data suitable for MetaHuman model manipulation, we contribute an emotional audio-visual dataset and derive control parameters for each frames. Our proposed methodology can be applied in producing dialogues animations of non-playable characters (NPCs) in video games, and driving avatars in virtual reality environments. Our further quantitative and qualitative experiments, as well as an user study comparing with existing researches show that our approach demonstrates superior results in driving 3D facial models. The code and sample data are available at \href{https://github.com/SJTU-Lucy/EmoFace}{https://github.com/SJTU-Lucy/EmoFace}
} 
\CCScatlist{
  \CCScatTwelve{Human-centered computing}{Computer graphics}{Graphics systems and interfaces}{Virtual reality}
}

\begin{document}


\firstsection{Introduction}

\maketitle
The ever-expanding development of virtual reality technology has led to a growing demand for the creation of virtual characters, and it has become an indispensable part in many domains. By creating avatars, we could put ourselves in a metaverse, and communicate via avatars. This mode of interaction empowers individuals to engage with others through avatar-mediated communication, thus bypassing the necessity of physical presence. It brings several advantages, including higher levels of anonymity and privacy, the opportunity to engage in virtual environments that might not be possible in the physical world. Furthermore, it has become a fundamental element of interactive technologies, finding applications in diverse fields. For instance, online virtual multiplayer games, social media platforms, virtual assistants, virtual meetings, and various other domains.\par

However, paradoxically, as the demand for realistic generated facial animations increases, people's tolerance for imperfections in the results diminishes, even in the case of subtle facial nuances. Even the smallest imperfection can induce the uncanny valley effect in the animated avatar, substantially decreasing its audience acceptance. \par

Traditionally, avatars can be generated through vision-based methods like face tracking, which have highly realistic outcomes. But a significant challenge arises when the user wears a headset, making the capture of facial expressions unfeasible. Under such circumstances, employing audio input as the foundation for generating avatars becomes a more suitable approach. Generally speaking, existing researches on audio-driven facial animation generation can be mainly concluded into three types: 

\paragraph{Video-based generation} Edits the video of the target character to achieve audio and video synchronization \cite{prajwal2020lip, zhou2021pose, zhou2019talking, song2018talking, chen2020talking};
\paragraph{Image-based generation} Uses one or several facial images as prototype for generation, and edited as a frame in the animation \cite{wang2021audio2head, lu2021live, ji2022eamm, wang2022one, saunders2023read};
\paragraph{Model-based generation} Uses controller rigs or facial mesh to drive the model or render facial animation \cite{edwards2016jali, taylor2017deep, richard2021audio, richard2021meshtalk, villanueva2022voice2face, fan2022faceformer}; 

Most previous studies focus on video-based and image-based generation, and few studies focus on model-based generation approaches. However, in terms of game production, it is more appropriate to use model-based approaches as the target characters appears in the form of 3D models.\par

The primary challenge in this task stems from the fact that speech audio includes more than just the phonemes of the spoken text. It also contains cues related to facial expressions. Consequently, a talking head should not only synchronize with the speech but also convey the speaker's emotional state through its expressions. While there have been notable successes in the research on audio-driven facial animation, the domain of multi-emotional generation has seen relatively limited exploration. Moreover, a significant proportion of existing datasets, such as MEAD \cite{wang2020mead}, are primarily based on English recordings, with an absence of datasets recorded in Chinese. Considering that we mainly use Chinese in application, and the substantial phonetic differences between Chinese and English, employing models trained on English data for Chinese audio clips can result in inaccurate facial animations. At the same time, the current datasets appear in the form of pairs of audio and video, and the complex mapping relationship between video and rig controller values to drive 3D models is hard to be learned. Consequently, the existing datasets cannot be directly utilized for model training. To address this issue, we propose an audio-visual dataset recorded in Chinese that contains seven different emotions. Through post-processing, we have extracted the controller values corresponding to each frames in the videos.\par

In addition to constructing the dataset, we also propose a fundamental face generation model tailored to this dataset,  which can be used for the facial generation in multiple emotions. This model takes an audio clip and the desired emotion as inputs, producing corresponding controller values for each frame to drive the MetaHuman model. However, given the relatively short duration of each recording, the dataset contains few instances of blinks and eye movements. Consequently, it becomes challenging to learn a robust correlation between blinks, eye gaze and speech, potentially leading to unnatural details in the generated talking head. To address this issue, we introduce independent blink and eye gaze control module. The blink controller gains blinking frequency data from other datasets and learns stochastic rules governing blinking behavior. Additionally, the gaze controller generates subtle eye movements, enhancing the naturalness of the facial animation.\par

This paper introduces EmoFace, a technology for driving virtual characters using audio and emotion as input. The mainstream researches can achieve good synchronization between input audios and output lip motions. However, these animations typically lack emotional expressions, with neutral face even when inputting emotional audio clips. Furthermore, the generated images from these methods are not suitable for driving virtual character models. To address these limitations, we propose an approach that takes both audio and emotional information as input and produces controller values for driving MetaHuman models, thereby enhancing the precision of facial animation generation. The main contributions of this paper are as follows:

\begin{itemize}
    \item We construct a dataset recorded in Chinese with multiple emotions, and extract the controller values of each frames;
    \item We propose a foundational model of audio-driven multi-emotional generation of MetaHuman controller rigs. This model offers the flexibility to control emotions and delivers high-quality facial animation;
    \item We enhance the facial expression generation process by incorporating blink and gaze controllers, thereby achieving a more natural and realistic outcome;
\end{itemize}

\section{Related Work}

\subsection{Audio-Driven Talking Face Generation}
The aim of audio-driven talking head generation is to produce an animation of the target character based on an audio clip while ensuring accurate lip synchronization. Existing research can be broadly categorized into three distinct types.\par

Some research works use GAN \cite{goodfellow2014generative} to directly output talking head videos. 
Song et al. \cite{song2018talking} introduced a novel conditional recurrent generation network. It incorporates reference images and audio signals into the recurrent unit to facilitate timing-dependent learning, thereby enhancing the temporal coherence of both images and audio signals. This enhancement ensures seamless transitions in lip and facial motion.
Vougioukas et al. \cite{vougioukas2019end, vougioukas2020realistic} employed a temporal generative adversarial network (TGAN). Within their approach, a generator features an encoder-decoder structure, combining raw audio and individual reference images, while a sequence discriminator is implemented to ensure the naturalness of the generated animations.
Wav2Lip \cite{prajwal2020lip} focuses on the synchronization of audio and mouth shape in face generation. It takes audio and a video with masked lower face as input. Then, a GAN is trained to fill the masked lower face, and a lip-sync loss, computed by SyncNet \cite{chung2017out}, is employed to guarantee the alignment of the generated faces with the audio.
The primary issue with GAN-based generation lies in its direct output of facial images for each frame, thus unable to be migrated to drive facial models.\par

Some studies focus on extracting phoneme from audio and learning the mapping between phoneme and viseme. 
JALI \cite{edwards2016jali} extracts phoneme sequence from text to animate the JALI rigs. Additionally, the system utilizes audio signal as an auxiliary input to predict the intensity of jaw and mouth movements based on audio features such as volume, pitch, and formant information.
Zhou et al. \cite{zhou2018visemenet} made enhancements to the JALI system by introducing a separation between the phoneme and landmark processes. Their approach involves the combination of phonemes, landmarks, and audio features to generate JA-LI parameter values and viseme information.
However, these approaches mainly focus on lip shapes and ignore the animation of other parts of the face. \par

Some researchers try to learn the mapping between audio and corresponding facial controller values. 
Pham et al. \cite{pham2017speech, pham2018end, pham2020learning} employed spectrograms as audio features, with the outcome being unit intensities corresponding to distinct facial regions. This is achieved by employing separate convolution processes in both the frequency and time domains.
VOCA \cite{cudeiro2019capture} utilizes a pre-defined character model and audio features extracted by DeepSpeech \cite{hannun2014deep} to generate facial meshes corresponding to the character.
FaceFormer \cite{fan2022faceformer} encodes the long-term audio context using the pretrained wav2vec2.0 \cite{baevski2020wav2vec} and employs a transformer decoder \cite{vaswani2017attention}, with carefully designed attention masks. This allows for the automatic regression and prediction of facial meshes.
MeshTalk \cite{richard2021meshtalk} aims to disentangle a latent space of facial animation through a classification process. In this disentangled space, audio-related information governs the lower face, while audio-irrelevant information influences the upper face.
While these methods have achieved a commendable level of authenticity in their results, they are constrained by the complex many-to-many mapping relationship that exists between audio and facial expressions. This direct mapping from audio to facial expressions may suffer from over-smoothing, indicating that the output tends to converge towards the mean value of similar examples within the dataset.\par

In the above researches, a notable limitation is the neglect of emotional expressions, resulting in outputs that lack emotion and being neutral.
Wang et al. \cite{wang2020mead} introduced the MEAD dataset, which aims to create emotional talking faces by independently segmenting the upper and lower parts of the face. Nevertheless, the animations generated using this approach did not have a high level of naturalness.
Building upon the groundwork by MEAD, Ji et al. propose EVP \cite{ji2021audio}, which extends the integration of emotions into the synthesis process. EVP utilizes time-aligned audio features in the form of MFCC (Mel-frequency cepstral coefficients) from the same text content under various emotional states during the training process. Moreover, EVP introduces a disentanglement module that effectively segregates content encoding from emotion encoding within the audio. EmoTalk \cite{peng2023emotalk} further implement wav2vec2.0 in the disentanglement module, making content and emotion further separated.\par

While most of the works focus on learning the mapping between audio and facial expressions,  FLINT \cite{danvevcek2023emotional} implements a VAE structure to learn facial motion priors. Although it could ease the problem of high-frequency jitter and other unnatural motions, details like blinks and gazes are ignored. Moreover, completing leaving aside audio could possibly cause lack of movements.

\subsection{Audio-Visual Dataset}
Currently there are some high-quality audio-visual datasets, but most of them do not consider emotional information. 
The LRW \cite{chung2017lip} is an automatically collected and processed compilation of data from British broadcast TV programs, offering a substantial diversity in its content. However, the dataset's inclusion of various roles and distinct individuals may introduce substantial interference, potentially hindering the model's ability to discern information that is irrelevant from individuals.
The VOCASET \cite{cudeiro2019capture} contains voice-face 4D scans from 6 female and 6 male subjects. 40 sentence fragments were collected for each subject, and the speech diversity was maximized. \par

A small number of datasets consider audio-visual data under multi-emotional conditions.
The dataset of Fanelli et al.\cite{fanelli20103} comprises a total of 1109 audio sequences, with an average duration of 4.67 seconds. Each participant contributed audio-4D facial scanning data of 40 spoken English sentences, and each sentence was recorded on two occasions: once with emotional expression and once without.
MEAD \cite{wang2020mead} is a multi-view, multi-emotional audio-visual dataset with different intensities. It is composed of 60 recorders with 8 different emotions. Each recorder within the dataset is associated with approximately 40 minutes of video content, providing a rich and diverse resource for research. 
EmoTalk \cite{peng2023emotalk} proposes a large-scale 3D emotional talking face (3D-ETF) dataset including both blendshape coefficients and mesh vertices. The dataset was based on two 2D audio-visual datasets: RAVDESS \cite{livingstone2018ryerson} and HDTF \cite{zhang2021flow}, and contains over 6.5 hours of data.\par

Sadly, existing audio-visual datasets have two problems. One is that the recorded audio is in English, which may have problems when extended to other languages. The other problem is that audio-visual data or blendshape coefficients are unsuitable for driving MetaHuman models.

\begin{figure*}
    \centering
    \includegraphics[width=2\columnwidth]{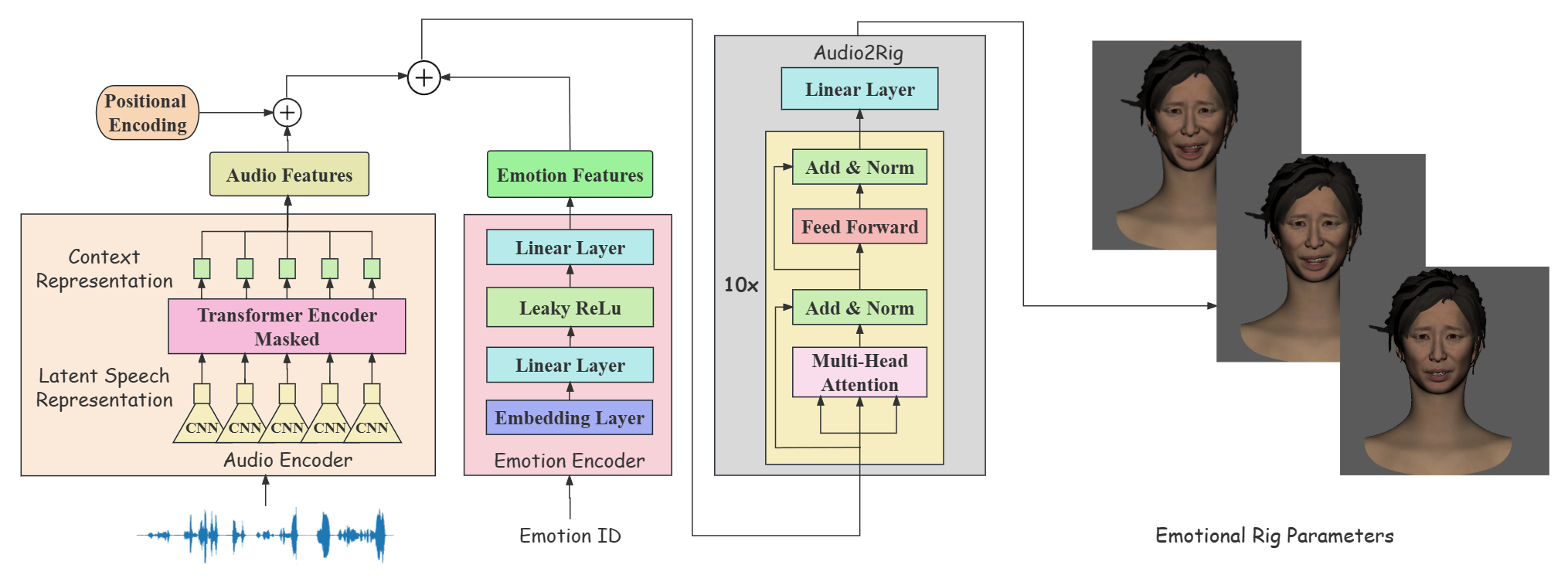}
    \caption{Structure of EmoFace}
    \label{fig:network}
\end{figure*}

\section{Method}
\subsection{Overview}
The architecture of our proposed model is illustrated in \autoref{fig:network}. The primary objective is to create an emotional, audio-driven talking head, while also enabling users to control the emotion and other details of the output facial animation. The model takes as input an audio clip and desired emotion, and yields the controller values required to drive the MetaHuman model to render the facial animation. It is composed of three parts, audio encoder, emotion encoder and Audio2Rig module. \par
\subsubsection{Audio Encoder}
Our audio encoder is constructed based on the self-supervised pre-trained speech model, wav2vec2.0 \cite{baevski2020wav2vec}. The multi-layer convolutional feature extractor takes raw audio as input and output latent speech representations with a frequency of 50. That is, the length of extracted features for 1s of audio would be 50. The encoder consists of several blocks containing temporal convolutional networks (TCN) \cite{bai2018empirical} followed by layer normalization and a GELU activation function. Then the output of the feature encoder is fed to a context network which follows the Transformer architecture to build representations from the entire sequence. \par

We initialize our audio encoder with a pre-trained wav2vec2.0 BASE model, which has been trained on a dataset of 960 hours of LibriSpeech data \cite{panayotov2015librispeech} sampled at a 16 kHz frequency. In this paper, we use wav2vec2.0 to extract the general features of the audio, and freeze the weights of the feature extractor throughout the training process. But the hidden states after feature extraction have a frame rate of 50, which is incompatible with our dataset recorded in frame rate of 60. To solve this, we implement a simple linear interpolation after this to ensure frequency alignment. And according to huggingface, attention mask should not be passed to wav2vec2-base structure to avoid degraded performance.\par

\subsubsection{Emotion Encoder}
In a corresponding manner, the emotion encoder accepts the emotion category from 0 to 6 as input and transforms it into a vector with identical dimensions to the content encoding. The emotion encoder is composed of an embedding layer and two fully connected layers. The embedding layer generates codes for different emotions, while the two fully connected layers further process these emotion-specific codes to produce encoded content with the same dimension. \par

As for the choice of input emotion, the model does not directly extract emotions from the audio, primarily because audio contains only a limited portion of emotion-related features. The emotion contained in facial expressions and the speech text is ignored, which is likely to cause inaccurate facial expressions. The audio-based extraction of emotions can be challenging. EVP \cite{ji2021audio} uses the MFCC of input audio clips to predict the emotion of input audio, but can only achieve an accuracy of 60\%. Meanwhile, EmoTalk \cite{peng2023emotalk} applies a model based on XLSR-Wav2Vec2 \cite{conneau2020unsupervised} for prediction. After fine-tuning on our dataset, it can achieve an accuracy of about 90\%. In addition, by inputting emotion label, users gain control over the emotional category for each frame, thus accurately obtain the required facial animation.

\subsubsection{Audio2Rig}
After collecting audio features and emotion encoding, they needs to be combined to form the input of Audio2Rig module. The audio features are first processed to produce content encoding. This content encoder comprises a fully connected layer and a positional encoding layer. The positional encoding layer serves to incorporate information regarding the relative position of tokens within the sequence window. The positional encoding has the same dimension as the model, and will be added to the input vector. In this regard, we use the original positional encoding in the transformer encoder, which has the form of sine and cosine functions of \autoref{con:sin} and \autoref{con:cos}. The \textit{pos} is the position in the input vector, \textit{i} is the dimension, \textit{$d_{model}$} is the feature dimension of the model, which is set to 512 in our model. By implementing sinusoidal version of positional encoding, model is able to extrapolate to sequence lengths longer than the ones encountered during training.

\begin{equation}
    PE_{(pos,2i)} = sin(\frac{pos}{10000^{\frac{2i}{d_{model} }  } } ) \label{con:sin}
\end{equation}
\begin{equation}
    PE_{(pos,2i+1)} = cos(\frac{pos}{10000^{\frac{2i}{d_{model} }  } } ) \label{con:cos}
\end{equation}

By combining the content encoding and emotion encoding, the result of 174-dimensional vector serves as input of Audio2Rig module. This module is composed of 10 transformer encoder layers and one fully connected layer to match dimensions for the output controller rigs. It is noteworthy that we depart from the approach employed in other studies, such as \cite{fan2022faceformer}, by deploying another Transformer encoder for prediction. Our choice to utilize a transformer encoder is mainly due to the following considerations:

\begin{itemize}
    \item Transformer encoder significantly outperforms Transformer decoder in terms of inference speed. We conducted inference time tests on short audio clips using both CPU and GPU. When on GPU, the average inference time is 21.57ms for Transformer encoder and 432.29ms for Transformer decoder. On the CPU, the times are 254.49ms for Transformer encoder and 811.48ms for Transformer decoder. This clearly demonstrates that employing a transformer encoder is better suited to meet real-time requirements.
    \item Transformer decoder traditionally incorporates all preceding values when predicting a new frame. However, in practice, only a few previous frames offer substantial insight for forecasting the next frame. Additionally, the wav2vec2 feature extractor already contains contextual information within the content encoding. Therefore, discarding previous frames has little impact on the result.
    \item To address potential jitter in the output sequence, we apply a Savitzky-Golay (savgol) filter with a window length of 15 frames and a polynomial order of 3. This filtering process ensures a smooth and visually pleasing facial animation.
\end{itemize}

\subsection{Blink}
While the above audio-driven model allows the generation of emotional talking heads with vivid expressions, a crucial issue remains: it lacks the action of blinks, which can significantly impact the user's perception of the generated animation.\par

Several factors contribute to the absence of blinks in the generated animations. Firstly, the dataset recordings are mainly short sentences less than 5 seconds in duration. Consequently, blink actions are rarely captured. Secondly, blinks do not exhibit a straightforward association with audio content. Other factors, such as head motions and intonation, can also influence blink frequency. As a result, it becomes challenging for the model to learn the underlying patterns related to blinks.\par

Previous study on blink rate \cite{bentivoglio1997analysis} highlights that blink rate is somewhat dependent on cognitive states, with tasks involving speaking or memory increasing blink rate and stable visual targets decreasing blink rate. Their results indicates that in the speaking state, the blink frequency, that is, the number of blinks in one minute, roughly conforms to the logarithmic normal distribution, with an average value of 26. Gender and age have no significant effect on blink frequency in this state. 
The research of Le et al.\cite{le2012live} indicates that blinking movements are mainly divided into involuntary blinking and voluntary eyelid movement. The former is usually completely closed, while the latter is usually related to head posture and emotional information. Their research also pointed out that the frequency of involuntary blinking basically conforms to the log-normal distribution law, in which the mean value of the frequency in the speaking mode is 21.1, and the standard deviation is 3.6.\par

\begin{figure}
    \centering
    \includegraphics[width=0.5\columnwidth]{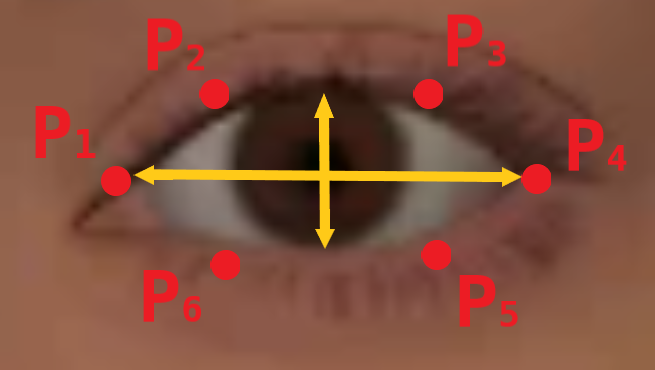}
    \caption{Positions of 6 landmarks for eye}
    \label{fig:eye_landmark}
\end{figure}

As for blink detection, we refer to the research of Cech \cite{cech2016real}. Existing research on predicting face landmarks \cite{xiong2013supervised, asthana2014incremental} typically utilize a single frame of static images as input and predict face landmarks using a set of 68 distinct points. Specifically, the left and right eyes are represented by 6 landmark points each, as illustrated in \autoref{fig:eye_landmark}. Once these landmarks have been acquired, the eye aspect ratio (EAR) can be computed according to \autoref{ear}. When the eyes are closed, the EAR value is approximately zero. The features of EAR includes qualities such as insensitivity to head pose, minimal variation between individuals, and resilience to uniform image scaling and rotation.

\begin{equation}
    EAR=\frac{||p_{2}-p_{6}||+ ||p_{3}-p_{5}|| }{2||p_{1}-p_{4}||}  \label{ear}
\end{equation}

A traditional approach for detecting eye blinks involves setting a threshold and considering a number of consecutive frames that fall below this threshold as a blink event. While this method is straightforward to implement, it carries the risk of misjudgment. Specifically, a low EAR value does not necessarily mean a blink event, as emotional expressions or certain facial movements can also cause a reduction in the EAR value, potentially leading to the error judge of blinking events. To address this issue, we explored the training of a support vector machine (SVM) classifier for blink detection using EAR values within a temporal sliding window. Given that the duration of blinking typically falls within the range of 0.1 to 0.4 seconds, and considering a video with a frame rate of 30 fps, we employ a sliding window of seven frames. This window includes the current frame, as well as three frames before and after. We create the training dataset by frame-by-frame labeling of videos.\par

\begin{figure}[h]
    \centering
    \includegraphics[width=\columnwidth]{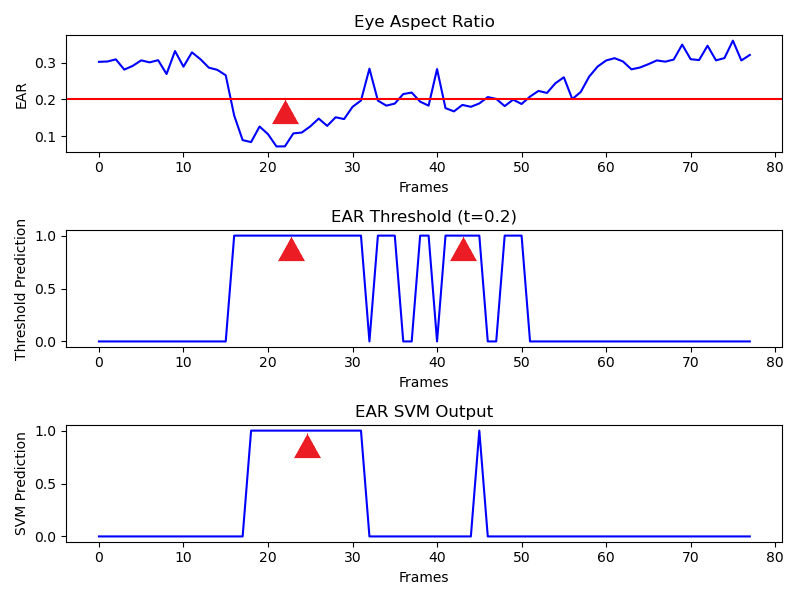}
    \caption{Comparison between SVM and threshold predictor}
    \label{fig:SVM}
\end{figure}

The method of using the EAR threshold to identify blinking is highly reliant on the precise selection of the threshold value. If the chosen value is too small, it may fail to recognize some blinks. Conversely, if set too high, actions induced by emotional expressions, such as squinting, may be classified as blinks, resulting in false recognition outcomes. \autoref{fig:SVM} presents a comparison between the results of the SVM predictor and the threshold-based predictor. Although there is one false-predicted frame by SVM predictor, as only consecutive frames will be considered as blink, the SVM prediction identifies one blink, whereas the threshold-based prediction identifies two. By employing the SVM model, which does not depend on exquisitely set EAR threshold, the occurrence of false blink recognition is effectively avoided, making the frequency statistics more reliable.\par

\begin{figure}
    \centering
    \includegraphics[width=0.9\columnwidth]{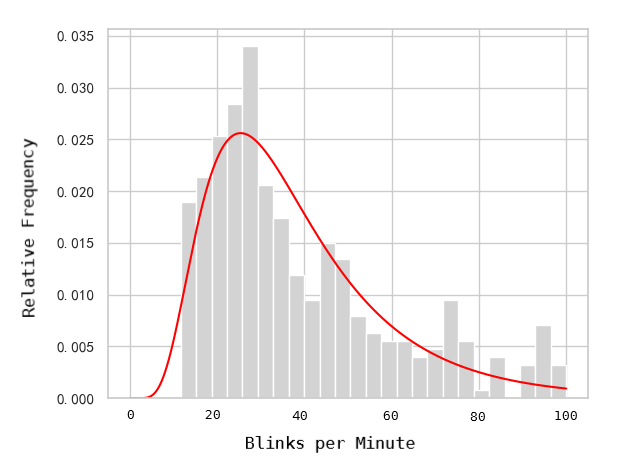}
    \caption{Collected blinks in a minute and its log-norm fitting curve}
    \label{fig:log-norm}
\end{figure}

Using the trained SVM model, we can collect the frequency of blinking from the dataset. Given that the videos are relatively short, we initially collect the time differences between two consecutive blinks and subsequently convert them into the number of blinks per minute for fitting purposes. \autoref{fig:log-norm} shows the collected frequency, and its fitting result of the log-norm distribution curve. Values exceeding 100 are excluded from the analysis. It can be seen from the results that the blink frequency basically satisfies the log-norm distribution. The mean and standard deviation of the natural logarithm of this distribution are calculated as 3.518 and 0.532, respectively.\par

With the log-normal curve successfully fitted, the time intervals between blinks can be sampled from this distribution. When it is time for a blink, the blinking action is regulated by multiplying the blinking control parameters generated by the original eye values within a window of 13 frames, considering that the target frame rate is 60 frames per second. The incorporation of an independent blink controller effectively governs the blinking action, ensuring that the avatar blinks at a plausible and randomized frequency. Our approach also provides users with the flexibility to freely adjust the blink rate by manipulating the distribution parameters.

\subsection{Gaze}
An additional issue related to upper face generation is the fixed eye gaze, which imparts a static and unnatural appearance to the talking head. This issue stems from the fact that the recording equipment is placed directly in front of the face. Consequently, during recording, the actor tends to maintain a straight-ahead gaze, leading to an absence of eye movement data in the dataset.\par

To tackle this issue, our initial approach involved extracting eye dynamics to train a model for audio-driven gaze prediction. We employed OpenFace \cite{amos2016openface} to capture the eye gaze angles from the MEAD dataset, forming a corresponding audio-gaze dataset. Subsequently, we deployed a model that combines wav2vec2.0 and a DNN to learn the mapping between audio and gaze. However, the result of this training reveals that the gaze actions generated by the model tends to remain static, with little variation in gaze angles. This can be due to the concentration of gaze angle data within a limited range, which causes the model's output towards the mean value of the training data. Then we doubled the gaze angles in the training dataset, which led to the emergence of eye darts. Nevertheless, a notable problem is that these eye darts demonstrate a high degree of consistency among different audio inputs and tend to favor a straight-ahead gaze over other directions.\par

Previous study has highlighted that eye movement is influenced by a multitude of factors, such as the accent of the spoken text and head posture \cite{le2012live}. Therefore, it is difficult in principle to learn the irregular mapping relationship between audio and gaze. Consequently, we turned our attention to Nvidia's Omniverse Audio2Face and observed its generated animations. While Audio2Face is capable of producing more natural blinks and eye movements, we observed a uniformity in facial movements across different audio inputs, similar to what we encountered with model-driven gaze generation. This uniformity, when applied to multiple audio inputs simultaneously, could result in a monotonous outcome. \par

\begin{figure}
    \centering
    \includegraphics[width=0.9\columnwidth]{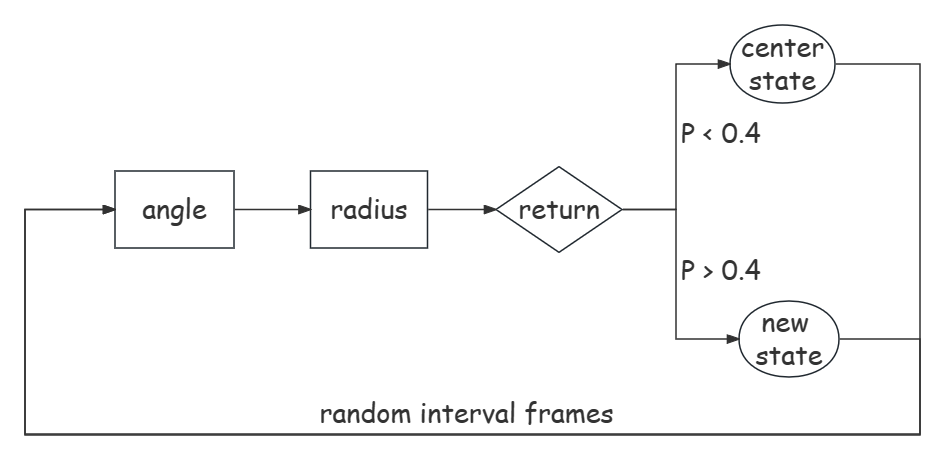}
    \caption{Gaze sampling procedure}
    \label{fig:gaze}
\end{figure}

Therefore, we considered a random yet efficient way. Parameters in the random generator include the range for sampling frame intervals, gaze radius and gaze angle, as shown in \autoref{fig:gaze}. In each cycle, the generator randomly samples the number of interval frames, radius and angle in the set range. And smoothly moves the eyeball to the angular position on the corresponding amplitude. To avoid constant movement, it has a 40\% chance of returning to the center of eye. This value is set by comparing between results of different possibilities. If lower, the frequency of eye rolls can be too high. If higher, there are cases of staring for a long time. By fine-tuning the range settings for the random sampler, we can achieve a more natural result and ensure compatibility with audio inputs of varying lengths. In this model, we have set the interval frame range to be between 15 and 45 frames (equivalent to 0.25-0.75 seconds), with a radius range of 0.1-0.2. The transitions between different states follow a linear interpolation approach.

\section{Experiment}
\subsection{Dataset}
Considering the absence of a suitable publicly available dataset, we undertook the task of recording and processing data to create our own audio-rig dataset. The data collection process involved gathering audio-visual data from a carefully chosen actor, who performed in seven different emotions. To ensure the accuracy and naturalness of the performances, we planned and executed the data collection process, including the choice of audio content, the method employed for data collection, and the post-processing of videos. Moreover, to guarantee the audio's quality, we thoughtfully selected emotionally consistent text covering possible phonemes. As different people have different ways to show emotions, the data collected from only one actor may limit the creativity of the model. We will expand the dataset by collecting data from different actors in the future.

For the design of the audio content, our aim was to maximize the phonemes within the speech material. The data for each emotion was divided into two segments: common and special. The common texts were devoid of explicit emotional cues and could be used across different emotions. In contrast, the special texts were tailored to convey specific emotional information and were exclusively employed in the dataset for that particular emotion. While for the emotion category, we defined seven different emotions including happy, sad, angry, surprised, fear, disgusted and neutral emotion states.\par

Upon recording the audio and videos, we proceeded to process this data to obtain the face controller values that serve as inputs for driving the MetaHuman model. A team of artists are responsible for converting video to rig sequence. They manually adjusted the controller values of the MetaHuman model to match the recorded video. By creating keyframes in this way, ground truth controller values can be obtained. The dataset consists of a total of 174 parameters. For each emotion, a distinct subset of these parameters is used, with any unused parameters set to 0. The 174 parameters separately control different parts of the face, mainly eye, jaw, mouth, teeth, tongue, brow, ear, nose and neck. Furthermore, the dataset was divided into a training set and a validation set, and the precise distribution is illustrated in \autoref{tab:data_distribution}.\par

\begin{table}[tb]
  \caption{Data distribution of different emotions}
  \label{tab:data_distribution}
  \scriptsize%
	\centering%
  \begin{tabu}{%
	r%
	*{7}{c}%
	*{2}{r}%
	}
  \toprule
   Emotion & \rotatebox{90}{Total} &   \rotatebox{90}{Neutral} &   \rotatebox{90}{Happy} &   \rotatebox{90}{Sad} &   
   \rotatebox{90}{Angry} &   \rotatebox{90}{Surprise} &   \rotatebox{90}{Fear} &   \rotatebox{90}{Disgusted} \\
  \midrule
  \textbf{Train} & 806 & 141 & 143 & 143 & 142 & 95 & 95 & 47 \\
  \textbf{Validation} & 41 & 7 & 7 & 7 & 7 & 5 & 5 & 3 \\
  \midrule
  \textbf{Total} & \textbf{847} & \textbf{148} & \textbf{150} & \textbf{150} & \textbf{149} & \textbf{100} & \textbf{100} & \textbf{50}\\
  \bottomrule
  \end{tabu}%
\end{table}

\begin{figure}
    \centering
    \includegraphics[width=1\columnwidth]{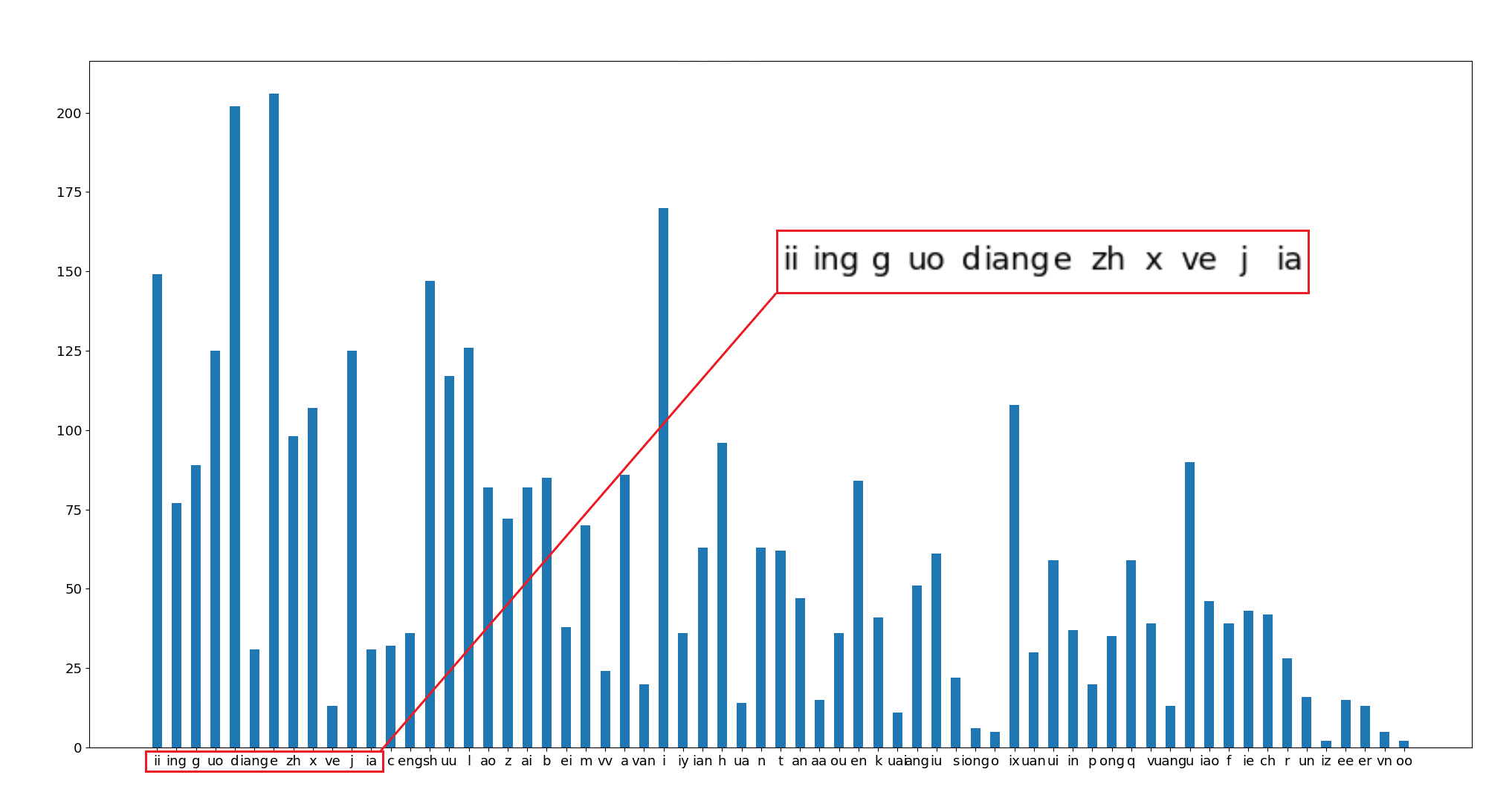}
    \caption{Phoneme distribution of happy emotion}
    \label{fig:phoneme}
\end{figure}

We used phoneme toolkit (phkit) perform to phoneme decomposition of the text, enabling the analysis of speech text content. Chinese phonemes differ from those in English as they include initials, finals, and tones. Our focus remained on the initials and finals, which comprise 27 types of initials and 41 types of finals. We tracked the number of occurrences of different phonemes, and phkit can decompose it into a total of 65 distinct phonemes. An example of phoneme distribution of happy emotion is shown in \autoref{fig:phoneme}. It is worth noting that the corpus we devised comprehensively covers nearly all phonemes, ensuring that the training set incorporates a wide range of mouth animations.\par

\begin{figure}
    \centering
    \includegraphics[width=\columnwidth]{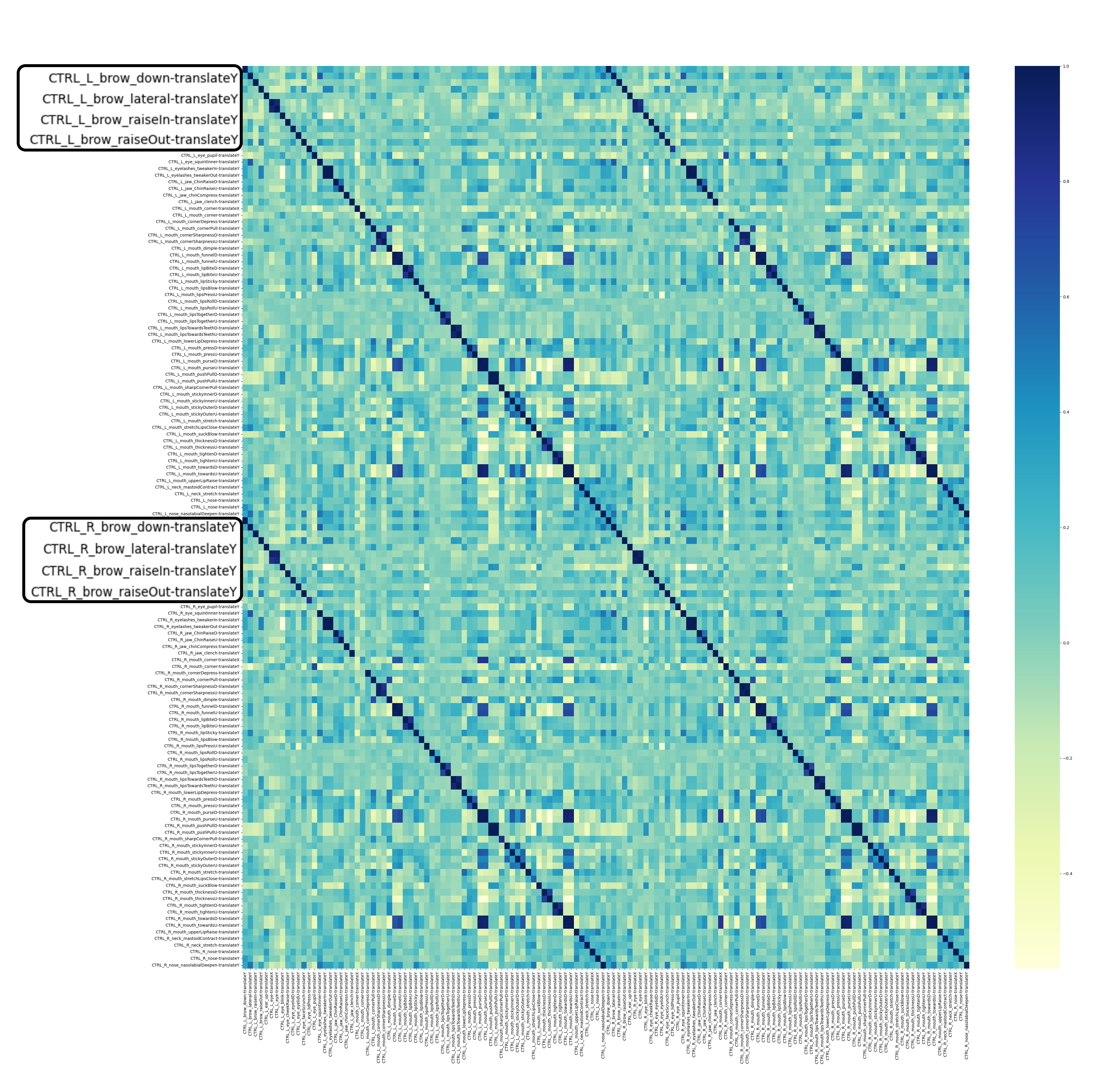}
    \caption{Correlation heat map of left and right face controller rigs}
    \label{fig:heatmap}
\end{figure}

Then, we conducted an analysis of the symmetry between the left and right part of the face. This involved the derivation of a correlation coefficient heat map that illustrates the relationships between parameters for the left and right sides of the face, as shown in \autoref{fig:heatmap}. As we can see, there are three diagonal lines in this heat map, which indicates that although the symmetric controller values are not strictly consistent, they do exhibit a high degree of correlation.

\subsection{Model Details}
In terms of the parameter settings, we employed the Adam optimizer. As for the learning rate strategy, training was conducted using a StepLR scheduler with a step size of 100 and a decay rate of 0.995. This means that every 100 rounds of training, the learning rate was reduced to 0.995 times its current value. In the selection of the loss function, we utilized the mean square error (MSE) to quantify the difference between the predicted 174 parameters for each frame and the ground truth. The training lasted a total of 3000 rounds.\par

Our model does not require audio pre-processing, the audio just needs to be loaded at a frequency of 16KHz. In the pre-training stage, we utilized the wav2vec2.0 BASE structure. Specifically, we employed the wav2vec2-base-960h pre-trained model. The weight of feature extractor in this pre-training model is frozen during training. The hidden states output by the pre-trained model is first mapped through a fully connected layer, and then the 512-dimensional vector is generated through the positional encoding layer. In the case of the emotion encoder, it comprises an embedding layer and two fully connected layers. We applied Leaky ReLU with a slope of 0.2 as the activation function between the fully connected layers. The emotion encoder also outputs a 512-dimensional encoding. The Audio2Rig module is structured as 10 transformer encoder layers and a fully connected layer. This component is responsible for mapping the 512-dimensional hidden states into a 174-dimensional controller rig sequence.

\section{Evaluation}
\subsection{Comparison to state-of-the-art}
We conducted a comparative analysis between the results produced by EmoFace, FaceFormer \cite{fan2022faceformer} and EmoTalk \cite{peng2023emotalk}. But the other two methods are designed for other datasets. FaceFormer uses VOCASET, an audio-mesh dataset. And EmoTalk proposed 3D-ETF based on RAVDESS \cite{livingstone2018ryerson} and HDTF \cite{zhang2021flow}, which is audio-blend shape correlated. To better compare them in our dataset, we modified FaceFormer and EmoTalk to fit the emotion label and output dimension of our dataset. We removed "template" in FaceFormer, "level" and "person" in EmoTalk. And the output dimensions for both models are set to 174.

\begin{table}[tbh]
  \caption{Comparison of MAE with state-of-the-art}
  \label{tab:sota}
  \scriptsize%
	\centering%
  \begin{tabu}{%
	r%
	*{7}{c}%
	*{2}{r}%
	}
  \toprule
   Model & \rotatebox{90}{Full-Face} &   \rotatebox{90}{Mouth-Area} &   \rotatebox{90}{Eye-Area} \\
  \midrule
  \textbf{EmoFace} & 0.04024 & 0.03697 & 0.05698 \\
  \textbf{FaceFormer} & 0.04114 & 0.03919 & 0.05389 \\
  \textbf{EmoTalk} & 0.04273 & 0.03963 & 0.05956 \\
  \midrule
  \end{tabu}%
\end{table}

\paragraph{Quantitative analysis}
We use the mean absolute value (MAE) of the prediction results and the ground truth to evaluate the models. The comparison is shown in \autoref{tab:sota}. The table indicates that EmoFace exhibits a smaller MAE, and its predictions are closer to the ground truth. However, the difference between EmoFace and FaceFormer is not significant. As EmoTalk requires paired audios as inputs, special audios that contained only in certain emotions cannot be used, which may have negative impact on the performance. What's more, EmoTalk does not directly use emotion label as input, but disentangles emotion embedding from the audio, any wrong prediction can seriously affect the predicted facial expression. To gain further insights, we conducted a separate analysis of the parameters in mouth area and eye area respectively. The controller rigs of the mouth region have a more significant impact on audio-lip synchronization. Conversely, the controller rigs affecting the eye area play a crucial role in conveying emotions. Our results demonstrate that EmoFace outperforms FaceFormer and EmoTalk in terms of mouth-related areas, although it slightly lags behind in other facial regions.

\begin{figure}
    \centering
    \includegraphics[width=0.7\columnwidth]{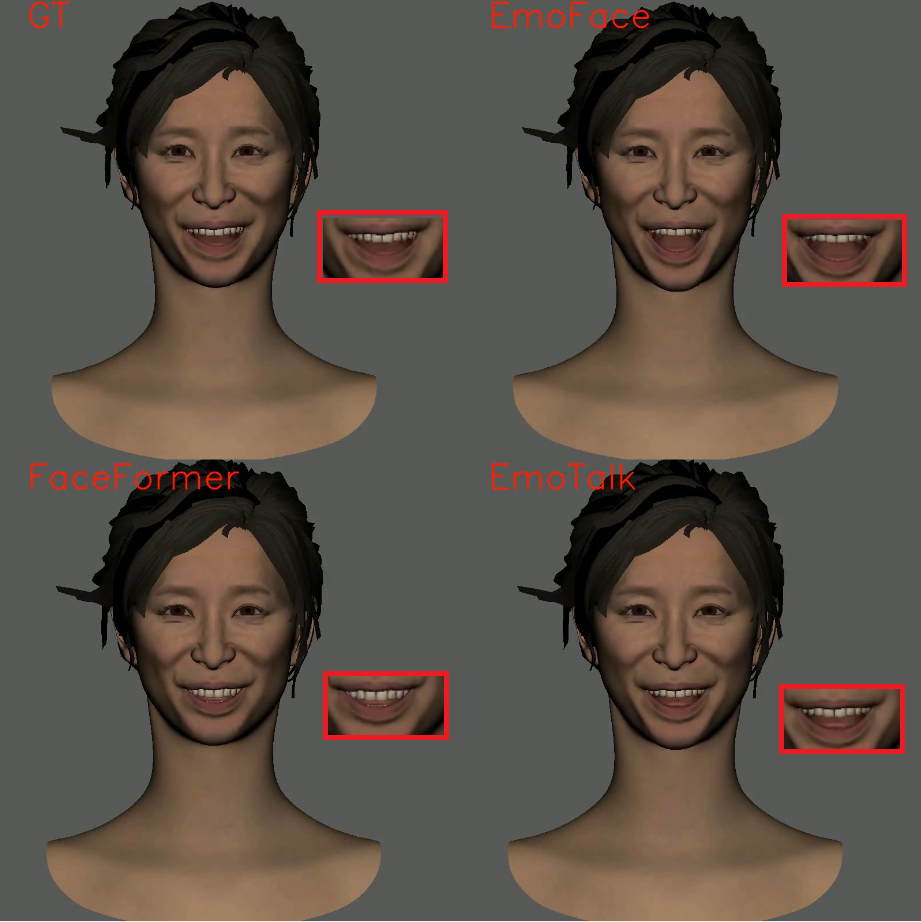}
    \caption{Comparison of rendered animation with state-of-the-art}
    \label{fig:sota}
\end{figure}

\paragraph{Qualitative analysis}
Considering the complex many-to-many mapping between upper face movements and audios, the quantitative evaluation metrics may not provide a completely accurate reflection of the prediction results. To address this, we rendered the predicted animations using a MetaHuman model, and the comparison is depicted in \autoref{fig:sota}. In terms of conveying emotional expressions, EmoFace and FaceFormer excel at capturing emotional characteristics. But as EmoTalk first predicts emotion from the audio clip, a wrong prediction can cause severe error in facial expression. However, when examining mouth dynamics, the lip movement of FaceFormer is slightly insufficient, resulting in half-open mouth in many cases. In contrast, the lip movements generated by EmoFace and EmoTalk are able to maintain synchronization with the audio.

\subsection{User Study}
We designed a user study to compare with FaceFormer and EmoTalk to evaluate our model.

\subsubsection{Participant and Design}
We recruited 15 participants from Shanghai Jiao Tong University for the user study, the average age was 22, ranging from 18 to 25 years old; 10 were men. They were naive to the purpose of the experiment. \par
 
We used the three trained models to render the two video clips for each emotion in a MetaHuman model for three different methods, as well as the extracted ground truth (GT).  \par

The experiment involved 7 emotions (Angry, Disgust, Fear, Happy, Neutral, Sad \& Surprise), 4 methods (GT, EmoFace, FaceFormer \& EmoTalk) and 2 sentences (Common sentences, Emotion-related sentences) in a within-subject design regarding emotions, and methods. The video clips were presented to the participants in random order. Each participant took part in 56 trials to evaluate the human expression, there were 840 trials in total.

\subsubsection{Procedure}
 Each participant were asked to answer the following questions for each rendered animation:
\begin{itemize}
    \item Emotion Recognition: ``Which emotion is expressed?" Participants were asked to select one emotion from: Angry, Disgust, Fear, Happy, Neutral, Sad and Surprise.
    \item Naturalness: ``How natural is the generated face?" Participants rated naturalness from 1 to 7, where 1 represents "not natural at all", and 7 represents “very realistic”.
    \item Lip Synchronization: ``Is the lip motion in sync with audio?" Participants rated on quality of lip synchronization from 1 to 7, where 1 represents not synchronized at all, and 7 represents perfect synchronization.
\end{itemize}

The whole experiment took about 15 minutes. The participants were paid 20 RMB amount. The experiment was approved by Shanghai Jiao Tong University Research Ethics Committee. \par

\begin{figure*}
    \begin{minipage}{0.32\textwidth}
        \centering
        \includegraphics[width=\linewidth]{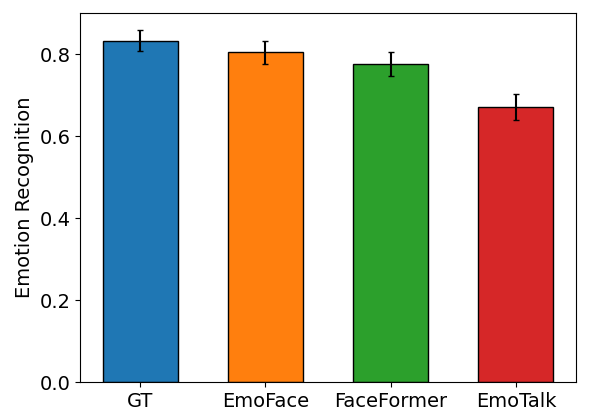}
    \end{minipage}
    \hfill
    \begin{minipage}{0.32\textwidth}
        \centering
        \includegraphics[width=\linewidth]{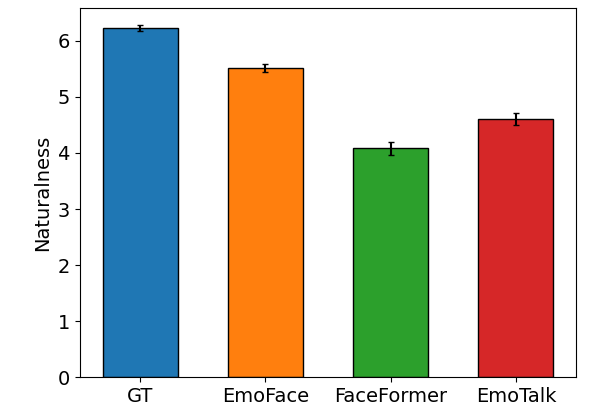}
    \end{minipage}
    \hfill
    \begin{minipage}{0.32\textwidth}
        \centering
        \includegraphics[width=\linewidth]{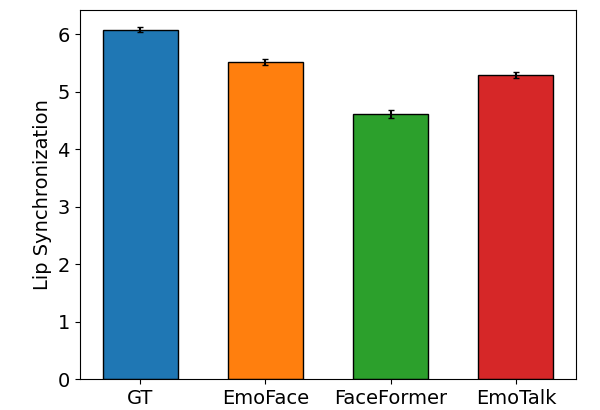}
    \end{minipage}
    \caption{Results of User Study}
    \label{fig:user_study}
\end{figure*}

\begin{table}[tbh]
  \caption{Comparison of ratings with 95\% confidence interval}
  \label{tab:user_study}
  \scriptsize%
	\centering%
  \begin{tabu}{%
	r% 
	*{7}{c}%
	*{2}{r}%
	}
  \toprule
   Model & \rotatebox{90}{Recognition} &   \rotatebox{90}{Naturalness} &   \rotatebox{90}{Lip-Sync} \\
  \midrule
  \textbf{GT} & \(0.833\pm0.051\) & \(6.228\pm0.098\) & \(6.076\pm0.083\) \\
  \textbf{EmoFace} & \(0.805\pm0.054\) & \(5.514\pm0.141\) & \(5.509\pm0.100\) \\
  \textbf{FaceFormer} & \(0.776\pm0.057\) & \(4.081\pm0.227\) & \(4.609\pm0.128\) \\
  \textbf{EmoTalk} & \(0.671\pm0.064\) & \(4.601\pm0.212\) & \(5.295\pm0.102\) \\
  \midrule
  \end{tabu}%
\end{table}

\subsubsection{Result}
After gaining rating from participants, we conducted separate repeated measures Analysis of Variances (ANOVAs). We calculated the average score for each method with an error bar of 95\% confidence interval, as shown in \autoref{tab:user_study}. We ran Mauchly’s test for validating sphericity, and when it is significant, we will apply Greenhouse-Geisser correction and mark the corrected result with “*”. Post-hoc tests were conducted using the Tukey test for the comparison of means.

\paragraph{Emotion Recognition} For the recognition of emotions, the choices of users are converted to 0 (incorrect) or 1 (correct). It can be seen that all three methods can well express emotion features. In contrast, our method achieves the best emotion recognition with average score of 0.809. FaceFormer also achieve 0.782, which is almost as good. However, the correctness for EmoTalk is significantly lower with 0.681. We used ANOVA to compare Faceformer and Emotalk respectively with our model. Results revealed the main effect of different models are significant with $ p=0.004^* $. The post-hoc result shows that our model is significantly better than EmoTalk with $ p=0.0059 $, but shows no significance to FaceFormer with $ p=0.8959 $. The result of emotion recognition shows that our method can precisely express emotion in the output facial animation.

\paragraph{Naturalness} The score for naturalness reflects the overall result of the generated facial animation. It can seen that our model with added blink and gaze achieves a rating of 5.559, which significantly surpasses the other two. While FaceFormer achieves 4.127 and EmoTalk achieves 4.569. Our method can achieve a score close to GT of 6.235. As for significance, our model is significantly better than FaceFormer and EmoTalk, with $ p<0.001^* $. The post-hoc result shows that our model is significantly better than EmoTalk and EmoFace with $ p<0.001 $ for both of them. It shows that our method can produce more natural facial expressions than the other two models.

\paragraph{Lip Synchronization} The rating for lip synchronization mainly reflects the accuracy of the lower half of the face. With pretrained wav2vec2 audio feature extractor, all three models can generate good lip synchronization with the audio. EmoFace (5.509) and EmoTalk (5.274) gain similar ratings, while FaceFormer (4.617) slightly lags behind. The main effect of the generation method is also significant, with $ p<0.001^* $. The result of post-hoc shows that our model is significantly better than EmoTalk and FaceFormer with $ p<0.001 $. It demonstrates that our model could produce more precise lip movements.

\subsection{Ablation study}
We conducted modifications to the EmoFace model to investigate the influence of different components on the prediction results. These modifications were divided into three categories: with and without weight initialization of the audio encoder, removing positional encoding, and the use of alternative structures to replace the transformer encoder in the Audio2Rig module. Each of the modified models was trained independently, and we subsequently performed quantitative analyses on the prediction results. We also apply MAE to evaluate these models. The outcomes of these analyses are presented in \autoref{tab:ablation}.

\begin{table}[tbh]
  \caption{Comparison between modified models}
  \label{tab:ablation}
  \scriptsize%
	\centering%
  \begin{tabu}{%
	r% 
	*{7}{c}%
	*{2}{r}%
	}
  \toprule
   Model & \rotatebox{90}{Full-Face} &   \rotatebox{90}{Mouth-Area} &   \rotatebox{90}{Eye-Area} \\
  \midrule
  \textbf{EmoFace} & 0.04024 & 0.03697 & 0.05698 \\
  \textbf{wo weight initialize} & 0.04706 & 0.04641 & 0.05826 \\
  \textbf{fc decoder} & 0.04166 & 0.03897 & 0.05703 \\
  \textbf{lstm decoder} & 0.04128 & 0.03862 & 0.05631 \\
  \textbf{wo positional encoding} & 0.04043 & 0.03743 & 0.05648 \\
  \midrule
  \end{tabu}%
\end{table}

\begin{figure}[h]
    \centering
    \includegraphics[width=\columnwidth]{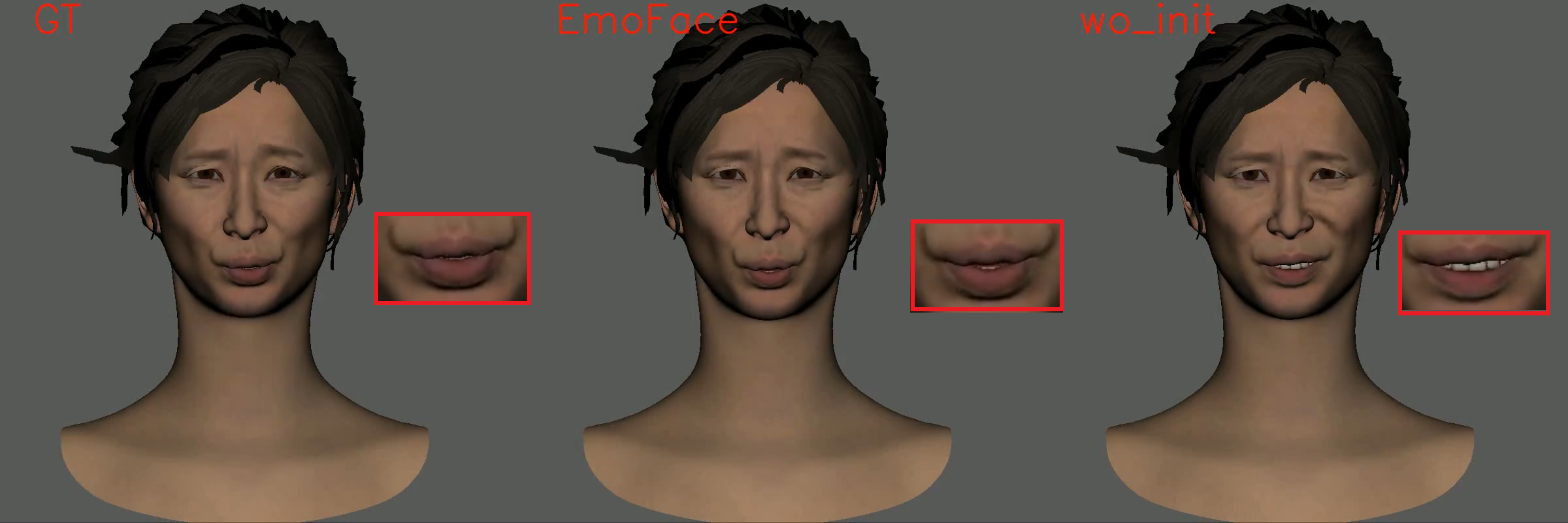}
    \caption{Comparison to EmoFace without initialization}
    \label{fig:init}
\end{figure}

\paragraph{Wav2vec2.0 initialization}
We conducted a comparison between the predicted animations with and without weight initialization. The results shows a clear deterioration in the quality of facial motion when weight initialization is not employed. Furthermore, the MAE value of the predictions increases significantly. Relatively speaking, the decline in the model's performance without the use of initialized weights, is mainly attributed to increased errors in mouth shape predictions. This observation highlights the importance of initializing with wav2vec2.0 pre-trained model. The prediction animations also exhibit issues such as unsynchronized audio and lip movements, as depicted in \autoref{fig:init}.

\begin{figure}[h]
    \centering
    \includegraphics[width=\columnwidth]{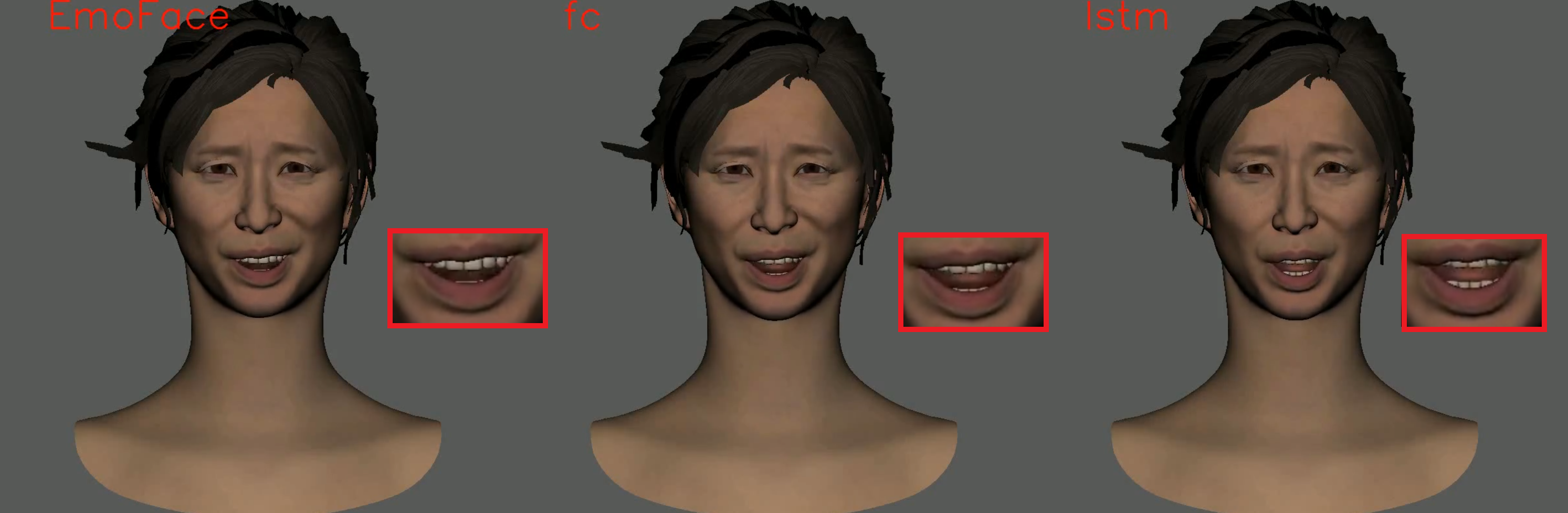}
    \caption{Comparison between different decoders}
    \label{fig:decoder}
\end{figure}

\paragraph{Decoder}
In the EmoFace model, we employed a 10-layer transformer encoder as the foundation for constructing the decoder. For comparison, we conducted additional training and testing with two distinct decoder architectures, one utilizing fully connected layers and the other implementing LSTM layers. In terms of quantitative analysis, when compared to the original model, the MAE of the fully connected decoder and the LSTM decoder exhibited relatively similar values for both the mouth shape and other parts of the face. However, the predicted animations reveal that when the audio reaches sections with lower volume, both decoders struggle to accurately represent the mouth movements, as illustrated in \autoref{fig:decoder}. In comparison, the LSTM decoder, which can capture time-series information, yields slightly better performance, whereas the fully connected decoder tends to output mouth shapes resembling silence.

\begin{figure}[h]
    \centering
    \includegraphics[width=\columnwidth]{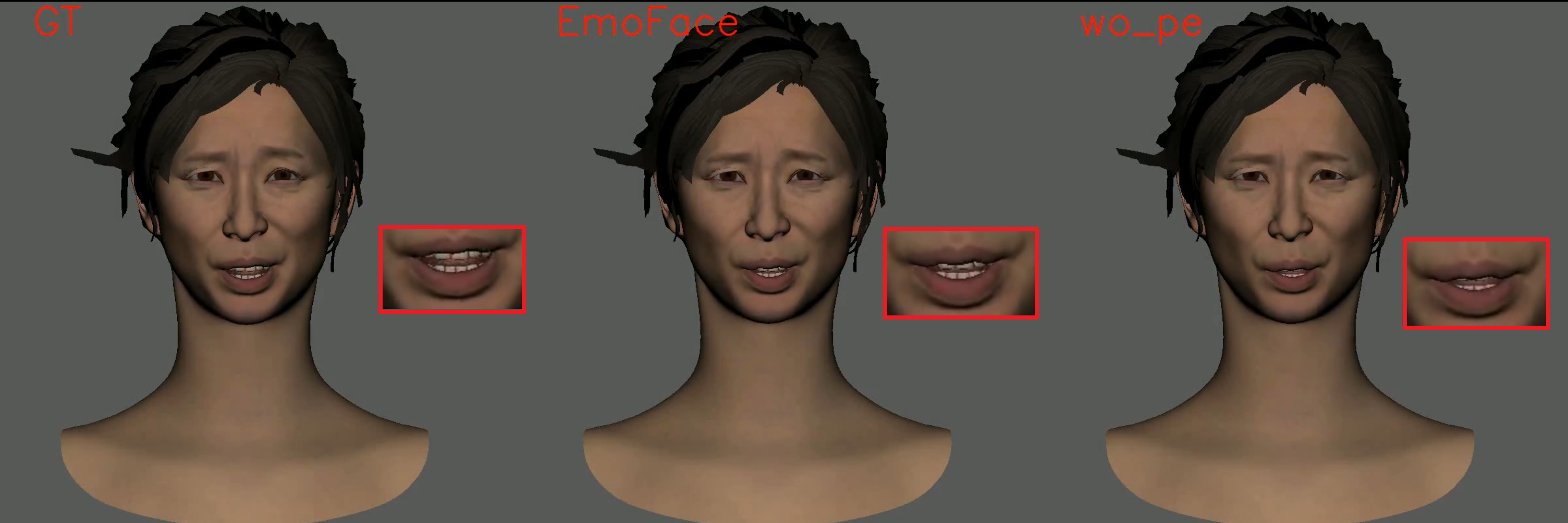}
    \caption{Comparison to EmoFace without positional encoding}
    \label{fig:pe}
\end{figure}

\paragraph{Positional Encoding}
The position encoding strategy we employ involves adding a temporal bias to the hidden feature and incorporating positional information into the hidden states. As a comparison, we attempted to remove the positional encoding component from the model to assess its impact on the model's output results. Upon the removal of positional encoding, the MAE of the predictions did not show significant changes. Judging from the generated animation, it became evident that the removal of positional encoding had little overall impact. This is likely because contextual information is already embedded in the features extracted by the content encoder. However, the absence of relative position information from positional encoding can lead to some inaccuracies in predictions, especially when rapid changes occur in the mouth shape, as illustrated in \autoref{fig:pe}.

\section{Discussion}
While our method can already generate realistic emotional facial animations, there are some limitations that need to be addressed. First, our method does not entirely resolve the challenge of multiple mappings between audio and facial expressions, potentially resulting in a lack of fine-grained detail in other facial areas. This issue is currently managed by introducing separate controllers for specific parts of the face, but it may become more obvious when working with larger datasets. Second, our model is built upon a large pre-trained language model, which leads to longer inference times and may not be suitable for real-time applications. Third, our dataset consists of data from a single actor in Chinese, which has relative small scale. Not only does it limit the diversity of generated expressions, but the subjectivity ability of the actor directly affects the performance of the model. Fourth, the generated expressions still suffer from the problem of lacking emotional intensity and lack of facial details comparing with motion captured ones, making it unable to meet the needs of cutscene-level applications. Our future work can be divided into two parts, expanding our dataset with data of different characters, emotion intensities and even languages, and exploring better face generation model using latest architectures such as diffusion models.\par

Our proposed method has a wide range of applications. It can be used in various fields, including game and movie production, where we can efficiently generate target animations from audio clips. In traditional production processes, creating facial expression videos for expression transfer or manually adjusting model parameters for each animation frame can be time-consuming. In comparison, audio-driven generation methods offer significant advantages in terms of speed and efficiency.\par

Moreover, in avatar-mediated communication, audio-driven face animation can play a pivotal role by synchronizing the avatar's facial dynamics with the spoken words of its user. This method effectively tackles the challenge of missing facial expressions when wearing a VR headset. It capitalizes on voice recognition to detect subtle speech nuances, such as pitch, tone, and rhythm, and translates them into dynamic real-time facial animations. This innovation not only enhances the expressiveness and engagement in communication but also forges deeper emotional connections in virtual interactions, bridging the gap between the digital and physical worlds. As we continue to explore this remarkable technology, it holds the potential to elevate avatar-mediated communication into a transformative and indispensable tool within our ever-increasingly digitalized society.

\section{Conclusion}
In this work, we introduce a novel approach to generate multi-emotional 3D facial animations driven by audio input. Our model, EmoFace, employs a pre-trained audio encoder to extract essential audio features, which are then combined with emotion encoding to produce facial controller values through the Audio2Rig module. Additionally, we incorporate supplementary blink and eye gaze controllers into the system to ensure more lifelike results. To train this model, we propose an emotional audio-visual dataset and derive the controller rigs for each frame. In essence, EmoFace excels at the task of animating the MetaHuman model with emotional audio inputs, producing outcomes with superior lip synchronization and emotionally expressive facial expressions.

\acknowledgments{
This work was supported by National Natural Science Foundation of China
(NSFC, NO. 62102255), the Open Project Program of the State Key Laboratory of CAD\&CG (Grant No. A2305), Zhejiang University, CCF-Tencent Open 
Research Fund (RAGR20220128).}

\bibliographystyle{abbrv-doi}

\bibliography{template}
\end{document}